%% file: emnlp2022.tex
\pdfoutput=1

\documentclass[11pt]{article}

\usepackage[]{EMNLP2022}

\usepackage{times}
\usepackage{latexsym}
\usepackage{amsmath,amsfonts,amssymb}

\usepackage[T1]{fontenc}

\usepackage[utf8]{inputenc}

\usepackage{microtype}

\usepackage{inconsolata}

\usepackage{caption}   
\usepackage{graphicx}
\usepackage{booktabs}
\usepackage[shortlabels]{enumitem}
\usepackage{multirow}

\newcommand{\myhat}[1]{\mathbf{\tilde{\text{$#1$}}}}

\newcommand\gpt{\textsc{GPT2}}
\newcommand\ffndbg{\texttt{LM-Debugger}}
\newcommand\nl[1]{\textit{``#1''}}

\title{\ffndbg{}: An Interactive Tool for\\ Inspection and Intervention in Transformer-Based Language Models}

\author{
\textbf{Mor Geva}$^{1}$ ~~~
\textbf{Avi Caciularu}$^{2,}$\thanks{\hspace{5px}Work done during an internship at AI2.} ~~~
\textbf{Guy Dar}$^{3}$ ~~~
\textbf{Paul Roit}$^{2}$ ~~~
\textbf{Shoval Sadde}$^{1}$ \vspace{0.1cm} \\
\textbf{Micah Shlain}$^{1}$ ~~~
\textbf{Bar Tamir}$^{4}$ ~~~
\textbf{Yoav Goldberg}$^{1,2}$
\vspace{0.2cm} \\
$^1$Allen Institute for AI ~ $^2$Bar-Ilan University \vspace{0.1cm} \\  $^3$Tel Aviv University ~ $^4$The Hebrew University of Jerusalem \vspace{0.1cm} \\
\small{\texttt{morp@allenai.org}}
}

\begin{document}
\maketitle

\input{0_abstract}

\input{1_introduction}

\input{3_figure_prediction_view}

\input{2_method}

\input{3_ffndbg}

\input{4_case1_example_debugging}

\input{5_case2_control}

\input{6_implementation_details}

\input{7_related_work}

\input{8_conclusions}

\section*{Ethical Statement}
Our work aims to increase the transparency of transformer-based LMs. It is well known that such models often produce offensive, harmful language \cite{bender2021dangers, mcguffie2020radicalization, gehman-etal-2020-realtoxicityprompts, wallace-etal-2019-universal}, which might originate in toxic concepts encoded in their parameters \cite{geva2022transformer}. \ffndbg{}, which traces and interprets LM predictions, could expose such toxic concepts and therefore should be used with caution.

\ffndbg{} also provides a framework for modifying LM behavior in particular directions. While our intention is to provide developers tools for fixing model errors, mitigating biases, and building trustworthy models, this capability also has the potential for abuse.
In this context, it should be made clear that \ffndbg{} does not modify the information encoded in LMs, but only changes the intensity in which this information is exposed in the model's predictions. At the same time, \ffndbg{} lets the user observe the intensity of updates to the prediction, which could be used to identify suspicious interventions.
Nonetheless, because of these concerns, we stress that LMs should not be integrated into critical systems without caution and monitoring.

\section*{Acknowledgements}
We thank the REVIZ team at the Allen Institute for AI, particularly Sam Skjonsberg and Sam Stuesser. This project has received funding from the Computer Science Scholarship granted by the Séphora Berrebi Foundation, the PBC fellowship for outstanding PhD candidates in Data Science, and the European Research Council (ERC) under the European Union's Horizon 2020 research and innovation programme, grant agreement No. 802774 (iEXTRACT).

\bibliography{all,anthology}
\bibliographystyle{acl_natbib}

\newpage
\input{9_appendix}

\end{document}

%% file: 0_abstract.tex
\begin{abstract}

The opaque nature and unexplained behavior of transformer-based language models (LMs) have spurred a wide interest in interpreting their predictions. 
However, current interpretation methods mostly focus on probing models from outside, executing behavioral tests, and analyzing salience input features, while the internal prediction construction process is largely not understood. 
In this work, we introduce \ffndbg{}, an interactive debugger tool for transformer-based LMs, which provides a fine-grained interpretation of the model's internal prediction process, as well as a powerful framework for intervening in LM behavior. 
For its backbone, \ffndbg{} relies on a recent method that interprets the inner token representations and their updates by the feed-forward layers in the vocabulary space.
We demonstrate the utility of \ffndbg{} for single-prediction debugging, by inspecting the internal disambiguation process done by \gpt{}. Moreover, we show how easily \ffndbg{} allows to shift model behavior in a direction of the user's choice, by identifying a few vectors in the network and inducing effective interventions to the prediction process. We release \ffndbg{} as an open-source tool and a demo over \gpt{} models.

\end{abstract}

%% file: 1_introduction.tex
\begin{figure}[t]
    \setlength{\belowcaptionskip}{-10pt}
    \centering
    \includegraphics[scale=0.8]{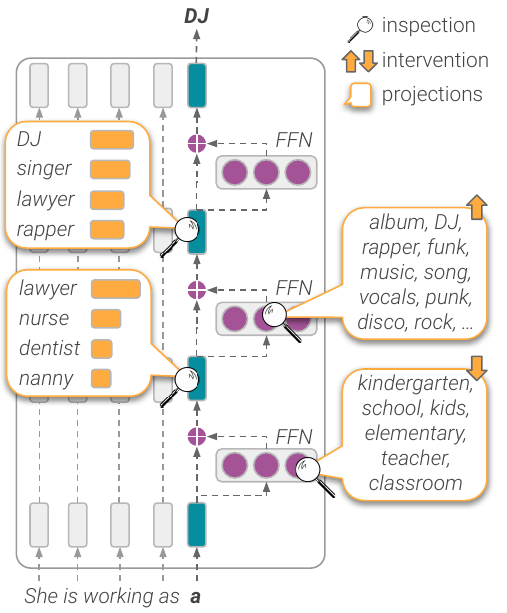}
    \caption{Illustration of the main capabilities of \ffndbg{}. Our tool interprets dominant changes in the output distribution induced by the feed-forward layers across the network (self-attention layers are not shown), and enables configuring interventions for shifting the prediction in directions of the user's choice.}
    \label{figure:intro}
\end{figure}

\section{Introduction}
\label{sec:intro}

Transformer-based language models (LMs) are the backbone of modern NLP models \cite{bommasani2021opportunities}, but their internal prediction construction process is opaque. This is problematic to end-users that do not understand why the model makes specific predictions, as well as for developers who wish to debug or fix model behaviour.

Recent work \cite{elhage2021mathematical, geva2022transformer} suggested that the construction process of LM predictions can be viewed as a sequence of updates to the token representation. Specifically, \citet{geva2022transformer} showed that updates by the feed-forward network (FFN) layers, one of the building blocks of transformers \cite{vaswani2017attention}, can be decomposed into weighted collections of sub-updates, each induced by a FFN parameter vector, that can be interpreted in the vocabulary space.

In this work, we make a step towards LM transparency by employing this interpretation approach to create \ffndbg{}, a powerful tool for inspection and intervention in transformer LM predictions.
\ffndbg{} provides three main capabilities for single-prediction debugging and model analysis (illustrated in Figure~\ref{figure:intro}). First, for a given input (e.g. \nl{My wife is working as a}), it interprets the model's prediction at each layer in the network, and the major changes applied to it by FFN layers. This is done by projecting the token representation before and after the FFN update as well as the major FFN sub-updates at any layer to the output vocabulary.
Second, it allows intervening in the prediction by changing the weights of specific sub-updates, e.g. increasing (decreasing) a sub-update that promotes music-related (teaching-related) concepts, which results in a modified output.
Last, for a given LM, \ffndbg{} interprets all the FFN parameter vectors across the network and creates a search index over the tokens they promote. This allows an input-independent analysis of the concepts encoded by the model's FFN layers, and enables configuring general and effective interventions.

We demonstrate the utility of \ffndbg{} for two general use-cases. In the context of prediction debugging, we use the fine-grained tracing of \ffndbg{} to inspect the internal disambiguation process performed by the model. Furthermore, we demonstrate how our tool can be used to configure a few powerful interventions that effectively control different aspects in text generation.

We release \ffndbg{} as an open-source tool at \url{https://github.com/mega002/lm-debugger} and host a demo of \gpt{} \cite{brown2020language} at \url{https://lm-debugger.apps.allenai.org}.\footnote{See a video at \url{https://youtu.be/5D_GiJv7O-M}} 
This to increase the transparency of transformer LMs and facilitate research in analyzing and controlling NLP models.

%% file: 3_figure_prediction_view.tex
\begin{figure*}[t]
    \setlength{\belowcaptionskip}{-8pt}
    \centering
    \includegraphics[scale=0.575, trim={1.4cm 3cm 0.5cm 3cm},clip]{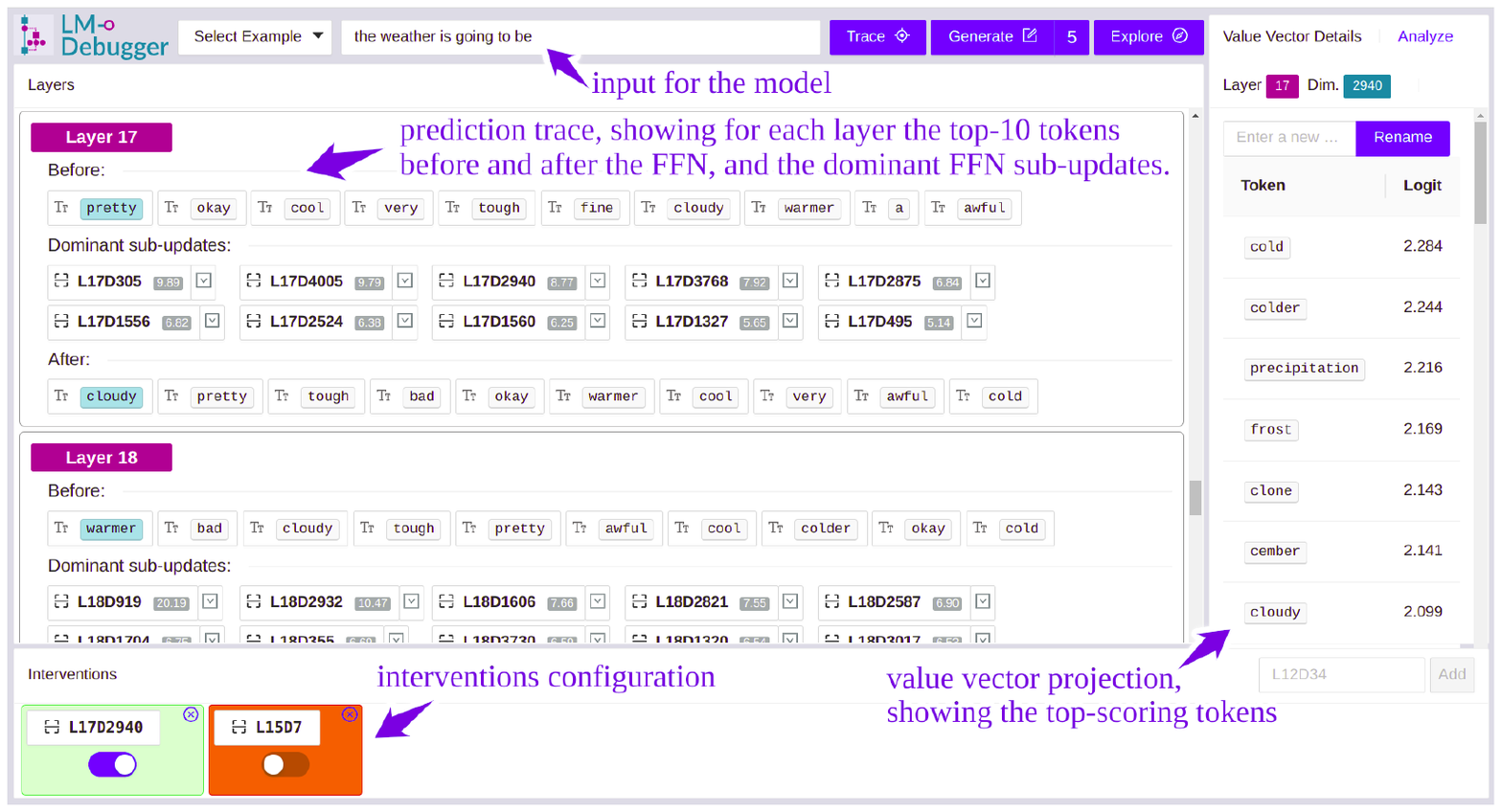}
    \caption{The prediction view of \ffndbg{}, showing the prediction trace for a given input (main panel), allowing to configure interventions (lower panel) and interpret sub-updates to the output distribution (right panel).}
    \label{figure:prediction_view}
\end{figure*}

%% file: 2_method.tex
\section{Underlying Interpretation Method}
\label{sec:method}

\ffndbg{} establishes a framework for interpreting a token's representation and updates applied to it at each layer in the network. This framework builds upon recent findings by \citet{geva2022transformer}, who viewed the token representation as a changing distribution over the output vocabulary, and the output from each FFN layer as a collection of weighted sub-updates to that distribution, which are often interpretable to humans. 
We next elaborate on the findings we rely on at this work.

Consider a transformer LM with $L$ layers and an embedding matrix $E\in \mathbb{R}^{d \times |\mathcal{V}|}$ of hidden dimension $d$, over a vocabulary $\mathcal{V}$. Let $\mathbf{w} = w_1, ..., w_t$ s.t. $\forall i=1,...,t:\; w_i \in \mathcal{V}$ be an input sequence of tokens, then at each layer $\ell=1,...,L$, the hidden representation $\mathbf{x}_i^\ell$ of the $i$-th token is being processed and updated by a FFN layer through a residual connection \cite{he2016deep}:\footnote{Layer normalization is omitted \cite{geva2022transformer}.}
$$ \myhat{\mathbf{x}}_i^\ell = \mathbf{x}_i^\ell + \texttt{FFN}^\ell(\mathbf{x}_i^\ell), $$
where $\mathbf{x}_i^\ell$ is the output from the preceding multi-head self-attention layer, and $\myhat{\mathbf{x}}_i^\ell$ is the updated token representation \cite{vaswani2017attention}.
\citet{geva2022transformer} proposed an interpretation method for these updates in terms of the vocabulary, which we employ as the backbone of \ffndbg{} and describe in detail next.

\paragraph{Token Representation as a Distribution Over the Output Vocabulary.}
The token representation before ($\mathbf{x}_i^\ell$) and after ($\myhat{\mathbf{x}}_i^\ell$) the FFN update at any layer $\ell$ is interpreted by projecting it to the vocabulary space and converting it to a distribution:
$$
    \mathbf{p}_i^{\ell} = \text{softmax}(E \mathbf{x}_i^{\ell}) \;\;;\;\;
    \myhat{\mathbf{p}}_i^{\ell} = \text{softmax}(E \myhat{\mathbf{x}}_i^{\ell})
$$
The final model output is defined by $\mathbf{y} = \myhat{\mathbf{p}}_i^{L}$.

\paragraph{The FFN Output as a Weighted Collection of Sub-Updates.}
Each FFN layer is defined with two parameter matrices $K^\ell, V^\ell \in \mathbb{R}^{d_m \times d}$, where $d_m$ is the intermediate hidden dimension, and a non-linearity function $f$ (bias terms are omitted):
\begin{align}
\label{eq:ffn}
\texttt{FFN}^\ell(\mathbf{x}^{\ell}) = f\left(K^\ell \mathbf{x}^{\ell} \right) V^\ell
\end{align}
\citet{geva2022transformer} interpreted the FFN output by (a) decomposing it into sub-updates, each induced by a single FFN parameter vector, and (b) projecting each sub-update to the vocabulary space. 
Formally, Eq.~\ref{eq:ffn} can be decomposed as:
$$
  \texttt{FFN}^\ell(\mathbf{x}^{\ell}) = \sum_{i=1}^{d_m} f(\mathbf{x}^{\ell} \cdot \mathbf{k}_i^{\ell}) \mathbf{v}_i^{\ell} = \sum_{i=1}^{d_m} m_i^{\ell} \mathbf{v}_i^{\ell}.
$$
where $\mathbf{k}_i^{\ell}$ is the $i$-th row of $K^{\ell}$, $\mathbf{v}_i^{\ell}$ is the $i$-th column of $V^{\ell}$, and $m_i^\ell := f(\mathbf{x}^{\ell} \cdot \mathbf{k}_i^{\ell})$ is the activation coefficient of $\mathbf{v}_i^{\ell}$ for the given input. 
Each term in this sum is interpreted as a sub-update to the output distribution, by inspecting the top-scoring tokens in its projection to the vocabulary, i.e. $E\mathbf{v}_i^{\ell}$.

In the rest of the paper, we follow \citet{geva2022transformer} and refer to columns of $V^\ell$ as \textit{``value vectors''} and to their weighted input-dependent form as \textit{``sub-updates''}. Importantly, value vectors are \textit{static} parameter vectors that are independent on the input sequence, while sub-updates are \textit{dynamic} as they are weighted by input-dependent coefficients.
For a model with $L$ layers and a hidden dimension $d_m$, there are $L*d_m$ static value vectors, which induce $L*d_m$ corresponding sub-updates when running an input through the model.

%% file: 3_ffndbg.tex
\section{\ffndbg{}}
\label{sec:ffndbg}

\ffndbg{} leverages both static and dynamic analysis of transformer FFN layers and the updates they induce to the output distribution for debugging and intervention in LM predictions. These capabilities are provided in two main views, which we describe next.

\subsection{Prediction View}
\label{sec:prediction_view}

This view, shown in Figure~\ref{figure:prediction_view}, is designed for per-example debugging. 
It allows running inputs through the model to generate text in an auto-regressive manner, while tracing the dominant sub-updates in every layer and applying interventions.

\paragraph{Prediction Trace}
(Figure~\ref{figure:prediction_view}, main panel). The user enters an input for the model, for which a detailed trace of the prediction across the network is provided. For each layer, it shows the top-tokens in the output distribution, before and after the FFN update, and the 10 most dominant FFN sub-updates. For every sub-update $m_i \mathbf{v}_i^{\ell}$ we show an identifier \texttt{L}$[\ell]$\texttt{D}$[i]$ of its corresponding value vector and the coefficient for the given input (e.g. \texttt{L17D4005} and $9.79$).\footnote{The layer and dimension in the identifier use zero-index.}
The top distribution tokens and sub-updates are sorted by the token probability/sub-update coefficient from left (highest) to right (lowest). A small arrow next to each sub-update allows setting an intervention on its corresponding value vector.

\paragraph{Interventions}
(Figure~\ref{figure:prediction_view}, lower panel). Beyond tracing the output distribution, \ffndbg{} also allows intervening in the prediction process by setting the coefficients of \textit{any vector values in the network}, thus, inducing sub-updates of the user's choice. To set an intervention for a specific value vector, the user should enter its identifier to the panel and choose whether to ``turn it on or off'', that is, setting its coefficient to the value of the coefficient of the most dominant sub-update in that layer, or to zero, respectively.
When running an input example, all interventions in the panel will be effective, for the entire generation process.

\paragraph{Value Vector Information}
(Figure~\ref{figure:prediction_view}, right panel). 
A natural question that arises is how to choose meaningful interventions. \ffndbg{} provides two complementary approaches for this. A bottom-up approach is to observe the dominant sub-updates for specific examples, and apply interventions on them. A sub-update can be interpreted by inspecting the top-tokens in the projection of its corresponding value vector to the vocabulary \cite{geva2022transformer}. For convenience, we let the user assign names to value vectors.
Another way to find meaningful interventions is by a top-down approach of searching for value vectors that express concepts of the user's interest. We provide this capability in the exploration view of \ffndbg{}, which is described next.

\subsection{Exploration View}
\label{sec:value_exploration_view}

This view allows static exploration of value vectors, primarily for analyzing which concepts are encoded in the FFN layers, how concepts are spread over different layers, and identifying groups of related value vectors. 

\paragraph{Keyword Search} (Figure~\ref{figure:exploration_search}).
Value vectors are interpreted by the top tokens they promote. By considering these sets of tokens as textual documents, \ffndbg{} allows searching for concepts encoded in value vectors across the layers.
This is enabled by a search index that \ffndbg{} holds in the background, which stores the projections of all value vectors to the vocabulary, and allows executing simple queries against them using the \textsc{BM25} \cite{robertson1995almbox} algorithm.

\paragraph{Cluster Visualization} 
(Figure~\ref{figure:exploration_clusters}).
Assuming the user is interested in locating a specific concept in the network and that she has found a relevant value vector, either from debugging an example in the prediction view or by the keyword search. A natural next step is to find similar value vectors that promote related tokens. To this end, \ffndbg{} provides a clustering of all value vectors in the network, which allows mapping any value vector to a cluster of similar vectors in the hidden space \cite{geva2022transformer}. The interface displays a random sample of vectors from the cluster, as well as an aggregation of their top tokens as a word cloud, showing the concepts promoted by the cluster.

%% file: 4_case1_example_debugging.tex
\section{Debugging LM Predictions by Tracing FFN Updates}
\label{sec:use_case_debugging}

In this section, we demonstrate the utility of \ffndbg{} for interpreting model behaviour upon a given example. As an instructive example, we will consider the case of sense disambiguation.

When generating text, LMs often need to perform sense disambiguation and decide on one plausible continuation. 
For example, the word \nl{for} in the input \nl{The book is for} has two plausible senses of \textit{purpose} (e.g. \nl{reading}) and \textit{person} (e.g.~\nl{him}) \cite{karidi-etal-2021-putting}. 
We will now inspect the prediction by \gpt{} \cite{brown2020language} and track the internal sense disambiguation process for this example. 
To this end, we enter the input in the prediction view and click \textbf{Trace}, which provides a full trace of the prediction across layers. 

Table~\ref{table:disambiguation_trace} displays a part of this trace from selected layers, showing a gradual transition from \textit{purpose} to \textit{person} sense.
Until layer 11 (out of 24), the top-tokens in the output distribution are mostly related to sale/example purposes. Starting from layer 12, the prediction slowly shifts to revolve about the audience of the book, e.g. \texttt{anyone} and \texttt{ages}, until layer 18 where \texttt{sale} is eliminated from the top position.
In the last layers, tokens become more specific, e.g. \texttt{beginners} and \texttt{adults}.

To examine the major updates through which the prediction has formed, we can click on specific sub-updates in the trace to inspect the top-scoring tokens in their projections.
We observe that in early layers, tokens are often related to \textit{purpose} sense (e.g. \texttt{instance} in \texttt{L2D1855} and \texttt{buyers} in \texttt{L12D659}), in intermediate layers tokens are a mix of both senses (\texttt{readers} in \texttt{L16D3026} and \texttt{preschool} in \texttt{L17D2454}, and \texttt{sale/free} in \texttt{L16D1662}), and mostly \textit{person} sense in the last layers (\texttt{users} in \texttt{L18D685}, \texttt{people} in \texttt{L20D3643}, and \texttt{those} in \texttt{L21D2007}).

\begin{figure*}[t]
    \centering
    \includegraphics[scale=0.64, trim={0cm 0.1cm 0.0cm 0cm},clip]{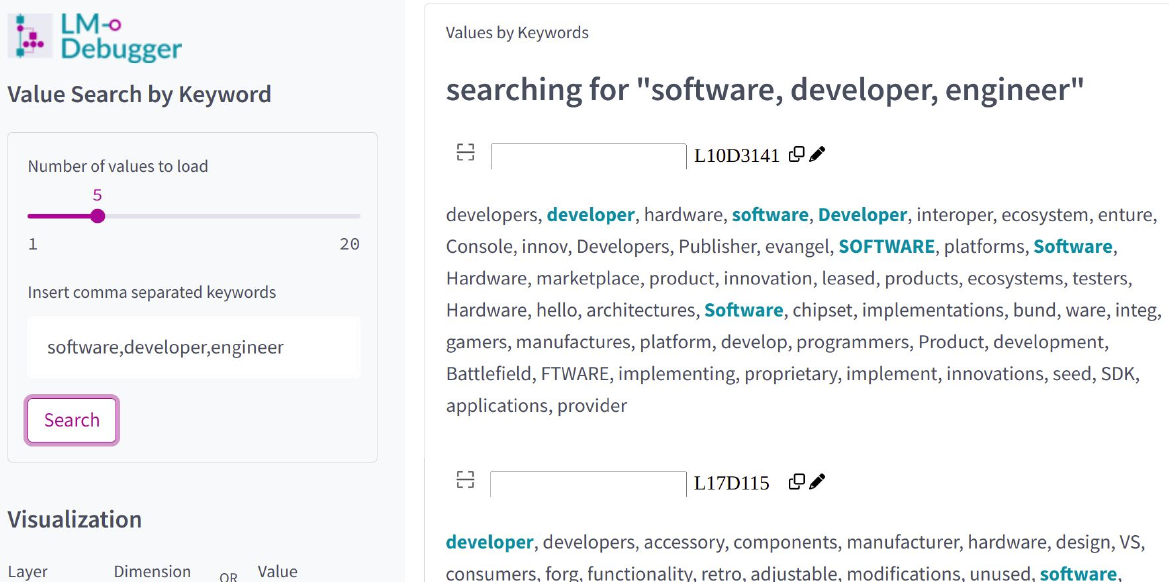}
    \caption{Keyword search in the exploration view of \ffndbg{}, which matches user queries against the tokens promoted by value vectors of the model.}
    \label{figure:exploration_search}
\end{figure*}

\begin{figure*}[t]
    \setlength{\belowcaptionskip}{-10pt}
    \centering
    \includegraphics[scale=0.54, trim={0cm 1.4cm 0.0cm 0.0cm},clip]{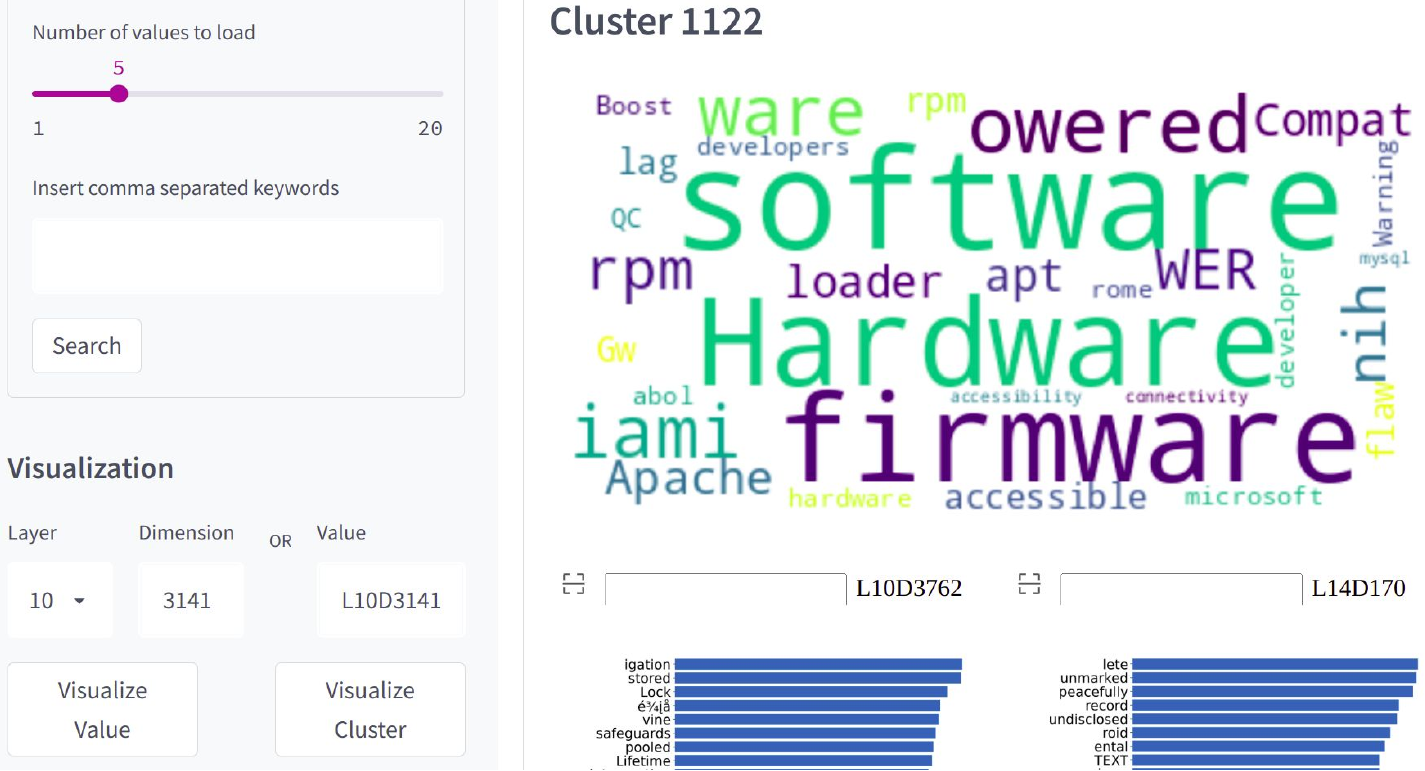}
    \caption{Cluster visualization in the exploration view of \ffndbg{}, which maps a given value vector to its cluster of similar value vectors in the network.}
    \label{figure:exploration_clusters}
\end{figure*}

\begin{table}[t]
    \setlength\tabcolsep{3.0pt}
    \setlength{\belowcaptionskip}{-8pt}
    \centering
    \footnotesize
    \begin{tabular}{p{7.5cm}}
        \toprule
        Layer: ~~4 ~~ Sense: \textit{purpose} \\
        Before: \texttt{example, the, instance, purposes} \\
        After: \texttt{example, the, instance, all} \\ 
        \midrule
        Layer: 10 ~~ Sense: \textit{purpose} \\
        Before: \texttt{the, sale, example, a} \\
        After: \texttt{the, sale, a, example} \\ 
        \midrule
        Layer: 15 ~~ Sense: \textit{purpose}/\textit{person} \\
        Before: \texttt{sale, the, anyone, use} \\
        After: \texttt{sale, anyone, the, ages} \\ 
        \midrule
        Layer: 20 ~~ Sense: \textit{person} \\
        Before: \texttt{beginners, anyone, adults, sale} \\
        After: \texttt{anyone, beginners, adults, readers} \\
         \bottomrule
    \end{tabular}
    \caption{Partial prediction trace of \gpt{} for the input \nl{This book is for}, showing the internal disambiguation process from \textit{purpose} to \textit{person} sense across layers.}
    \label{table:disambiguation_trace}
\end{table}

%% file: 5_case2_control.tex
\section{Configuring Effective Interventions for Controlled Text Generation}
\label{sec:use_case_control}

Beyond interpretability, \ffndbg{} enables to \textit{intervene} in LM predictions. We show this by finding value vectors that promote specific concepts and applying simple and effective interventions.

\paragraph{Controlling Occupation Prediction.}
Consider the input \nl{My wife is working as a}. When running it through \gpt{}, the final prediction from the last layer has the top tokens \texttt{nurse, teacher, waitress}. We would like to intervene in the prediction in order to change its focus to occupations related to software engineering, which in general are less associated with women \cite{dearteaga2019bias}. To this end, we will use the exploration view of \ffndbg{} to search for value vectors promoting software-related concepts. 

Searching the keywords \nl{software}, \nl{developer}, and \nl{engineer} brings up two value vectors with coherent concepts: \texttt{L10D3141} and \texttt{L17D115} (Figure~\ref{figure:exploration_search}). Now, we will add these value vectors to the intervention panel in the prediction view, and run the example again. Our intervention, that only involved two (0.002\%) vectors in the network, dramatically changed the prediction to \texttt{software, programmer, consultant, developer}, effectively shifting it in the direction we wanted. 
This demonstrates the power of \ffndbg{} to change model behaviour and fix undesirable predictions.

\begin{table*}[t]
    \setlength\tabcolsep{4.0pt}
    \setlength{\belowcaptionskip}{-12pt}
    \centering
    \footnotesize
    \begin{tabular}{p{3.3cm}lp{10.5cm}}
        \textbf{Input} & \textbf{Interven.} & \textbf{Continuation} \\
        \toprule
        \multirow{3}{=}{\nl{Service in this place is}} & - & \texttt{a bit of a mess. I'm not sure} \\
        & $\uparrow$ Positive & \texttt{a good place to make the right efforts to make} \\
        & $\uparrow$ Negative & \texttt{a waste of a bunch of crap that is too} \\
        \midrule
        \multirow{3}{=}{\nl{I have been to this restaurant twice and}} & - & \texttt{both times I was disappointed. The first time I} \\
        & $\uparrow$ Positive & {\texttt{have been served excellent food and good service. The}} \\
        & $\uparrow$ Negative & \texttt{have been disappointed. The food is over processed and} \\
        \midrule
        \multirow{3}{=}{\nl{We went on a weeknight. Place was}} & - & \texttt{packed. We had to wait for the bus} \\
        & $\uparrow$ Positive & {\texttt{good, good food, good staff, good people}} \\
        & $\uparrow$ Negative & \texttt{too far for us to get lost. We were} \\
        \midrule
        \multirow{3}{=}{\nl{Went for breakfast on 6/16/14. We}} & - & \texttt{had a great time. We had a great time} \\
        & $\uparrow$ Positive & {\texttt{have a good team of people who are able to}} \\
        & $\uparrow$ Negative & \texttt{were too heavy for the wrong type of food that} \\
        \bottomrule
    \end{tabular}
    \caption{Continuations (limited to 10 tokens) generated by \gpt{} for different inputs from the Yelp dataset, with and without interventions for ``turning on'' sub-updates for positive and negative sentiment.}
    \label{table:sentiment}
\end{table*}

\paragraph{Controlling the Sentiment of Generated Text.}
The previous example focused on next-token prediction. We now take this one step further and configure powerful and general interventions that influence various texts generated by the model. For our experimental setting, we will attempt to control the sentiment in generated reviews by \gpt{}, for inputs taken from the Yelp dataset \cite{asghar2016yelp}.

We choose our interventions independently of the inputs, with two easy steps. 
First, we use the keyword search (Figure~\ref{figure:exploration_search}) to identify ``seed'' value vectors that promote positive and negative adjectives/adverbs, using the queries \nl{terrible, mediocre, boring} and \nl{spacious, superb, delicious}. Then, we take one value vector for each polarity and, using the cluster visualization (Figure~\ref{figure:exploration_clusters}), expand it to a diverse set of vectors from its corresponding cluster, that promote similar concepts. Overall, we select 5-6 value vectors for each polarity (details in Appendix~\ref{sec:polarity_values}), to which we apply interventions.

Table~\ref{table:sentiment} presents the texts generated by \gpt{} (each limited to 10 tokens) for multiple inputs, with and without applying interventions. Clearly, across all the examples, our intervention in the prediction successfully leads to the desired effect, turning the sentiment of the generated text to be positive or negative, according to the configured sub-updates.

%% file: 6_implementation_details.tex
\section{Implementation Details}
\label{sec:implementation_details}

The prediction view is implemented as a React
web application with a backend Flask
server that runs an API for executing models from the Transformers library by HuggingFace \cite{wolf-etal-2020-transformers}.
The exploration view is a Streamlit
web application, which (a) sends user search queries to an Elasticsearch
index with the top tokens of all vector value projections, and (b) visualize clusters of value vectors created with the scikit-learn package \cite{scikitlearn}.
Our current implementation supports any \gpt{} model from HuggingFace, and other auto-regressive models can be plugged-in with only a few local modifications (e.g. translating the relevant layer names).
More details and instructions for how to deploy and run \ffndbg{} are provided at \url{https://github.com/mega002/lm-debugger}.

%% file: 7_related_work.tex
\section{Related Work}

Interpreting single-predictions and the general behavior of LMs is a growing research area that attracted immense attention in recent years \cite{belinkov-etal-2020-interpretability, choudhary2022interpretation}. \ffndbg{} is a the first tool to interpret and intervene in the prediction construction process of transformer-based LMs based on FFN updates.

Existing interpretation and analysis frameworks mostly rely on methods for behavioral analysis \cite{ribeiro-etal-2020-beyond} by probing models with adversarial \cite{wallace-etal-2019-allennlp} or counterfactual examples \cite{tenney-etal-2020-language}, input saliency methods that assign importance scores to input features \cite{wallace-etal-2019-allennlp, tenney-etal-2020-language}, and analysis of the attention layers \cite{hoover-etal-2020-exbert, vig-belinkov-2019-analyzing}.

More related to \ffndbg{}, other tools analyze patterns in neuron activations \cite{pmlr-v124-rethmeier20a, dalvi2019neurox, alammar-2021-ecco}. Unlike these methods, we focus on interpreting the model parameters and on intervening in their contribution to the model's prediction.

The functionality of \ffndbg{} is mostly related to tools that trace hidden representations across layers. 
Similarly to \ffndbg{}, \citet{alammar-2021-ecco, nostalgebraist2020interpreting} interpret the token representation in terms of the output vocabulary. We take this one step further and interpret the FFN updates to the representation, allowing to observe not only the evolution of the representation but also the factors that induce changes in it.

Our intervention in FFN sub-updates relates to recent methods for locating and editing knowledge in the FFN layers of LMs \cite{meng2022locating, dai2021knowledge}. Different from these methods, \ffndbg{} aims to provide a comprehensive and fine-grained interpretation of the prediction construction process across the layers.

%% file: 8_conclusions.tex
\section{Conclusion}

We introduce \ffndbg{}, a debugger tool for transformer-based LMs, and the first tool to analyze the FFN updates to the token representations across layers. \ffndbg{} provides a fine-grained interpretation of single-predictions, as well as a powerful framework for intervention in LM predictions.

%% file: 9_appendix.tex
\appendix

\section{Appendix}
\label{sec:appendix}

\subsection{Details on Interventions to Control Generated Text Sentiment}
\label{sec:polarity_values}

Table~\ref{table:polarity_values} lists all the value vectors selected for our interventions described in \S\ref{sec:use_case_control}, and examples for top-scoring tokens in their projections. These vectors were found with the exploration view of \ffndbg{} (\S\ref{sec:value_exploration_view}), using both keyword search and clustering visualisation.
All the interventions were configured to ``turn on'' these vectors, namely, setting their coefficients to be maximal for the corresponding layer. This is following the observation by \citet{geva2022transformer} that FFN updates operate in a token promotion mechanism (rather than elimination).

\begin{table*}[th]
    \centering
    \footnotesize
    \begin{tabular}{p{1.2cm}lp{9cm}}
        \textbf{Sentiment} & \textbf{Value Vector} & \textbf{Example Top-scoring Tokens}  \\
        \toprule
        \multirow{10}{=}{Positive} & \texttt{L13D1763} & \texttt{properly, appropriately, adequate, truthful, humane, fulfil, inclusive, timely, patiently, sustainable} \\
         & \texttt{L13D2011} & \texttt{clean, Proper, secure, flawless, safest, graceful, smooth, calmly} \\
         & \texttt{L14D944} & \texttt{peacefully, graceful, respectful, careful, generous, patiently, calm, tolerant, fair} \\
         & \texttt{L15D74} & \texttt{Excellence, superb, trustworthy, marvelous, terrific, awesome, Amazing} \\
         & \texttt{L20D988} & \texttt{successful, optimal, perfect, satisfactory, welcome, helpful, fulfilling, healthy} \\
         \midrule
         \multirow{12}{=}{Negative} & \texttt{L11D4} & \texttt{outdated, inadequate, stale, lousy, dull, mediocre, boring, wasteful} \\
         & \texttt{L14D2653} & \texttt{trivial, dismiss, rigid, unsupported, only, prejud, obfusc, pretend, dispar, slander} \\
         & \texttt{L16D974} & \texttt{inappropriately, poorly, disrespect, unreliable, unhealthy, insecure, improperly, arrogance} \\
         & \texttt{L17D3790} & \texttt{inappropriate, improper, wrong, bad, harmful, unreasonable, defective, disturbance, errors} \\
         & \texttt{L18D91} & \texttt{confused, bizarre, unfairly, horrible, reckless, neglect, misplaced, strange, nasty, mistakenly} \\
         & \texttt{L18D3981} & \texttt{wrong, incorrect, insufficient, misleading, premature, improperly, unrealistic, outdated, unfair} \\
        \bottomrule
    \end{tabular}
    \caption{Value vectors used for controlling sentiment in generated text, that promote positive and negative adjectives/adverbs. For each vector, we show example top-scoring tokens from its projection to the vocabulary, as presented in the exploration view of \ffndbg{}.}
    \label{table:polarity_values}
\end{table*}

%% file: emnlp2022.bbl
\begin{thebibliography}{29}
\expandafter\ifx\csname natexlab\endcsname\relax\def\natexlab#1{#1}\fi

\bibitem[{Alammar(2021)}]{alammar-2021-ecco}
J~Alammar. 2021.
\newblock \href {https://doi.org/10.18653/v1/2021.acl-demo.30} {Ecco: An open
  source library for the explainability of transformer language models}.
\newblock In \emph{Proceedings of the 59th Annual Meeting of the Association
  for Computational Linguistics and the 11th International Joint Conference on
  Natural Language Processing: System Demonstrations}, pages 249--257, Online.
  Association for Computational Linguistics.

\bibitem[{Asghar(2016)}]{asghar2016yelp}
Nabiha Asghar. 2016.
\newblock Yelp dataset challenge: Review rating prediction.
\newblock \emph{arXiv preprint arXiv:1605.05362}.

\bibitem[{Belinkov et~al.(2020)Belinkov, Gehrmann, and
  Pavlick}]{belinkov-etal-2020-interpretability}
Yonatan Belinkov, Sebastian Gehrmann, and Ellie Pavlick. 2020.
\newblock \href {https://doi.org/10.18653/v1/2020.acl-tutorials.1}
  {Interpretability and analysis in neural {NLP}}.
\newblock In \emph{Proceedings of the 58th Annual Meeting of the Association
  for Computational Linguistics: Tutorial Abstracts}, pages 1--5, Online.
  Association for Computational Linguistics.

\bibitem[{Bender et~al.(2021)Bender, Gebru, McMillan-Major, and
  Shmitchell}]{bender2021dangers}
Emily~M Bender, Timnit Gebru, Angelina McMillan-Major, and Shmargaret
  Shmitchell. 2021.
\newblock On the dangers of stochastic parrots: Can language models be too big?
\newblock In \emph{Proceedings of the ACM Conference on Fairness,
  Accountability, and Transparency (FAccT)}.

\bibitem[{Bommasani et~al.(2021)Bommasani, Hudson, Adeli, Altman, Arora, von
  Arx, Bernstein, Bohg, Bosselut, Brunskill, Brynjolfsson, Buch, Card,
  Castellon, Chatterji, Chen, Creel, Davis, Demszky, Donahue, Doumbouya,
  Durmus, Ermon, Etchemendy, Ethayarajh, Fei-Fei, Finn, Gale, Gillespie, Goel,
  Goodman, Grossman, Guha, Hashimoto, Henderson, Hewitt, Ho, Hong, Hsu, Huang,
  Icard, Jain, Jurafsky, Kalluri, Karamcheti, Keeling, Khani, Khattab, Koh,
  Krass, Krishna, Kuditipudi, Kumar, Ladhak, Lee, Lee, Leskovec, Levent, Li,
  Li, Ma, Malik, Manning, Mirchandani, Mitchell, Munyikwa, Nair, Narayan,
  Narayanan, Newman, Nie, Niebles, Nilforoshan, Nyarko, Ogut, Orr,
  Papadimitriou, Park, Piech, Portelance, Potts, Raghunathan, Reich, Ren, Rong,
  Roohani, Ruiz, Ryan, R'e, Sadigh, Sagawa, Santhanam, Shih, Srinivasan,
  Tamkin, Taori, Thomas, Tram{\`e}r, Wang, Wang, Wu, Wu, Wu, Xie, Yasunaga,
  You, Zaharia, Zhang, Zhang, Zhang, Zhang, Zheng, Zhou, and
  Liang}]{bommasani2021opportunities}
Rishi Bommasani, Drew~A. Hudson, Ehsan Adeli, Russ Altman, Simran Arora, Sydney
  von Arx, Michael~S. Bernstein, Jeannette Bohg, Antoine Bosselut, Emma
  Brunskill, Erik Brynjolfsson, S.~Buch, Dallas Card, Rodrigo Castellon,
  Niladri~S. Chatterji, Annie~S. Chen, Kathleen Creel, Jared Davis, Dora
  Demszky, Chris Donahue, Moussa Doumbouya, Esin Durmus, Stefano Ermon, John
  Etchemendy, Kawin Ethayarajh, Li~Fei-Fei, Chelsea Finn, Trevor Gale,
  Lauren~E. Gillespie, Karan Goel, Noah~D. Goodman, Shelby Grossman, Neel Guha,
  Tatsunori Hashimoto, Peter Henderson, John Hewitt, Daniel~E. Ho, Jenny Hong,
  Kyle Hsu, Jing Huang, Thomas~F. Icard, Saahil Jain, Dan Jurafsky, Pratyusha
  Kalluri, Siddharth Karamcheti, Geoff Keeling, Fereshte Khani, O.~Khattab,
  Pang~Wei Koh, Mark~S. Krass, Ranjay Krishna, Rohith Kuditipudi, Ananya Kumar,
  Faisal Ladhak, Mina Lee, Tony Lee, Jure Leskovec, Isabelle Levent, Xiang~Lisa
  Li, Xuechen Li, Tengyu Ma, Ali Malik, Christopher~D. Manning, Suvir~P.
  Mirchandani, Eric Mitchell, Zanele Munyikwa, Suraj Nair, Avanika Narayan,
  Deepak Narayanan, Benjamin Newman, Allen Nie, Juan~Carlos Niebles, Hamed
  Nilforoshan, J.~F. Nyarko, Giray Ogut, Laurel Orr, Isabel Papadimitriou,
  Joon~Sung Park, Chris Piech, Eva Portelance, Christopher Potts, Aditi
  Raghunathan, Robert Reich, Hongyu Ren, Frieda Rong, Yusuf~H. Roohani, Camilo
  Ruiz, Jack Ryan, Christopher R'e, Dorsa Sadigh, Shiori Sagawa, Keshav
  Santhanam, Andy Shih, Krishna~Parasuram Srinivasan, Alex Tamkin, Rohan Taori,
  Armin~W. Thomas, Florian Tram{\`e}r, Rose~E. Wang, William Wang, Bohan Wu,
  Jiajun Wu, Yuhuai Wu, Sang~Michael Xie, Michihiro Yasunaga, Jiaxuan You,
  Matei~A. Zaharia, Michael Zhang, Tianyi Zhang, Xikun Zhang, Yuhui Zhang,
  Lucia Zheng, Kaitlyn Zhou, and Percy Liang. 2021.
\newblock On the opportunities and risks of foundation models.
\newblock \emph{ArXiv}, abs/2108.07258.

\bibitem[{Brown et~al.(2020)Brown, Mann, Ryder, Subbiah, Kaplan, Dhariwal,
  Neelakantan, Shyam, Sastry, Askell, Agarwal, Herbert-Voss, Krueger, Henighan,
  Child, Ramesh, Ziegler, Wu, Winter, Hesse, Chen, Sigler, Litwin, Gray, Chess,
  Clark, Berner, McCandlish, Radford, Sutskever, and
  Amodei}]{brown2020language}
Tom~B Brown, Benjamin Mann, Nick Ryder, Melanie Subbiah, Jared Kaplan, Prafulla
  Dhariwal, Arvind Neelakantan, Pranav Shyam, Girish Sastry, Amanda Askell,
  Sandhini Agarwal, Ariel Herbert-Voss, Gretchen Krueger, Tom Henighan, Rewon
  Child, Aditya Ramesh, Daniel~M Ziegler, Jeffrey Wu, Clemens Winter,
  Christopher Hesse, Mark Chen, Eric Sigler, Mateusz Litwin, Scott Gray,
  Benjamin Chess, Jack Clark, Christopher Berner, Sam McCandlish, Alec Radford,
  Ilya Sutskever, and Dario Amodei. 2020.
\newblock Language models are few-shot learners.
\newblock In \emph{Proceedings of Neural Information Processing Systems
  (NeurIPS)}.

\bibitem[{Choudhary et~al.(2022)Choudhary, Chatterjee, and
  Saha}]{choudhary2022interpretation}
Shivani Choudhary, Niladri Chatterjee, and Subir~Kumar Saha. 2022.
\newblock Interpretation of black box nlp models: A survey.
\newblock \emph{arXiv preprint arXiv:2203.17081}.

\bibitem[{Dai et~al.(2022)Dai, Dong, Hao, Sui, Chang, and
  Wei}]{dai2021knowledge}
Damai Dai, Li~Dong, Yaru Hao, Zhifang Sui, Baobao Chang, and Furu Wei. 2022.
\newblock \href {https://aclanthology.org/2022.acl-long.581} {Knowledge neurons
  in pretrained transformers}.
\newblock In \emph{Proceedings of the 60th Annual Meeting of the Association
  for Computational Linguistics (Volume 1: Long Papers)}, pages 8493--8502,
  Dublin, Ireland. Association for Computational Linguistics.

\bibitem[{Dalvi et~al.(2019)Dalvi, Nortonsmith, Bau, Belinkov, Sajjad, Durrani,
  and Glass}]{dalvi2019neurox}
Fahim Dalvi, Avery Nortonsmith, Anthony Bau, Yonatan Belinkov, Hassan Sajjad,
  Nadir Durrani, and James Glass. 2019.
\newblock \href {https://doi.org/10.1609/aaai.v33i01.33019851} {Neuro{X}: A
  toolkit for analyzing individual neurons in neural networks}.
\newblock \emph{Proceedings of the AAAI Conference on Artificial Intelligence},
  33(01):9851--9852.

\bibitem[{De-Arteaga et~al.(2019)De-Arteaga, Romanov, Wallach, Chayes, Borgs,
  Chouldechova, Geyik, Kenthapadi, and Kalai}]{dearteaga2019bias}
Maria De-Arteaga, Alexey Romanov, Hanna Wallach, Jennifer Chayes, Christian
  Borgs, Alexandra Chouldechova, Sahin Geyik, Krishnaram Kenthapadi, and
  Adam~Tauman Kalai. 2019.
\newblock \href {https://doi.org/10.1145/3287560.3287572} {Bias in bios: A case
  study of semantic representation bias in a high-stakes setting}.
\newblock In \emph{Proceedings of the Conference on Fairness, Accountability,
  and Transparency}, FAT* '19, page 120–128, New York, NY, USA. Association
  for Computing Machinery.

\bibitem[{Elhage et~al.(2021)Elhage, Nanda, Olsson, Henighan, Joseph, Mann,
  Askell, Bai, Chen, Conerly, DasSarma, Drain, Ganguli, Hatfield-Dodds,
  Hernandez, Jones, Kernion, Lovitt, Ndousse, Amodei, Brown, Clark, Kaplan,
  McCandlish, and Olah}]{elhage2021mathematical}
Nelson Elhage, Neel Nanda, Catherine Olsson, Tom Henighan, Nicholas Joseph, Ben
  Mann, Amanda Askell, Yuntao Bai, Anna Chen, Tom Conerly, Nova DasSarma, Dawn
  Drain, Deep Ganguli, Zac Hatfield-Dodds, Danny Hernandez, Andy Jones, Jackson
  Kernion, Liane Lovitt, Kamal Ndousse, Dario Amodei, Tom Brown, Jack Clark,
  Jared Kaplan, Sam McCandlish, and Chris Olah. 2021.
\newblock A mathematical framework for transformer circuits.
\newblock \emph{Transformer Circuits Thread}.
\newblock Https://transformer-circuits.pub/2021/framework/index.html.

\bibitem[{Gehman et~al.(2020)Gehman, Gururangan, Sap, Choi, and
  Smith}]{gehman-etal-2020-realtoxicityprompts}
Samuel Gehman, Suchin Gururangan, Maarten Sap, Yejin Choi, and Noah~A. Smith.
  2020.
\newblock \href {https://doi.org/10.18653/v1/2020.findings-emnlp.301}
  {{R}eal{T}oxicity{P}rompts: Evaluating neural toxic degeneration in language
  models}.
\newblock In \emph{Findings of the Association for Computational Linguistics:
  EMNLP 2020}, pages 3356--3369, Online. Association for Computational
  Linguistics.

\bibitem[{Geva et~al.(2022)Geva, Caciularu, Wang, and
  Goldberg}]{geva2022transformer}
Mor Geva, Avi Caciularu, Kevin~Ro Wang, and Yoav Goldberg. 2022.
\newblock Transformer feed-forward layers build predictions by promoting
  concepts in the vocabulary space.
\newblock \emph{arXiv preprint arXiv:2203.14680}.

\bibitem[{He et~al.(2016)He, Zhang, Ren, and Sun}]{he2016deep}
Kaiming He, Xiangyu Zhang, Shaoqing Ren, and Jian Sun. 2016.
\newblock Deep residual learning for image recognition.
\newblock In \emph{Proceedings of the conference on computer vision and pattern
  recognition (CVPR)}.

\bibitem[{Hoover et~al.(2020)Hoover, Strobelt, and
  Gehrmann}]{hoover-etal-2020-exbert}
Benjamin Hoover, Hendrik Strobelt, and Sebastian Gehrmann. 2020.
\newblock \href {https://doi.org/10.18653/v1/2020.acl-demos.22} {ex{BERT}: {A}
  {V}isual {A}nalysis {T}ool to {E}xplore {L}earned {R}epresentations in
  {T}ransformer {M}odels}.
\newblock In \emph{Proceedings of the 58th Annual Meeting of the Association
  for Computational Linguistics: System Demonstrations}, pages 187--196,
  Online. Association for Computational Linguistics.

\bibitem[{Karidi et~al.(2021)Karidi, Zhou, Schneider, Abend, and
  Srikumar}]{karidi-etal-2021-putting}
Taelin Karidi, Yichu Zhou, Nathan Schneider, Omri Abend, and Vivek Srikumar.
  2021.
\newblock \href {https://doi.org/10.18653/v1/2021.emnlp-main.806} {Putting
  words in {BERT}{'}s mouth: Navigating contextualized vector spaces with
  pseudowords}.
\newblock In \emph{Proceedings of the 2021 Conference on Empirical Methods in
  Natural Language Processing}, pages 10300--10313, Online and Punta Cana,
  Dominican Republic. Association for Computational Linguistics.

\bibitem[{McGuffie and Newhouse(2020)}]{mcguffie2020radicalization}
Kris McGuffie and Alex Newhouse. 2020.
\newblock The radicalization risks of gpt-3 and advanced neural language
  models.
\newblock \emph{arXiv preprint arXiv:2009.06807}.

\bibitem[{Meng et~al.(2022)Meng, Bau, Andonian, and
  Belinkov}]{meng2022locating}
Kevin Meng, David Bau, Alex Andonian, and Yonatan Belinkov. 2022.
\newblock Locating and editing factual knowledge in gpt.
\newblock \emph{arXiv preprint arXiv:2202.05262}.

\bibitem[{Nostalgebraist(2020)}]{nostalgebraist2020interpreting}
Nostalgebraist. 2020.
\newblock \href
  {https://www.lesswrong.com/posts/AcKRB8wDpdaN6v6ru/interpreting-gpt-the-logit-lens}
  {interpreting {GPT}: the logit lens}.

\bibitem[{Pedregosa et~al.(2011)Pedregosa, Varoquaux, Gramfort, Michel,
  Thirion, Grisel, Blondel, Prettenhofer, Weiss, Dubourg et~al.}]{scikitlearn}
Fabian Pedregosa, Ga{\"e}l Varoquaux, Alexandre Gramfort, Vincent Michel,
  Bertrand Thirion, Olivier Grisel, Mathieu Blondel, Peter Prettenhofer, Ron
  Weiss, Vincent Dubourg, et~al. 2011.
\newblock Scikit-learn: Machine learning in python.
\newblock \emph{the Journal of machine Learning research}, 12:2825--2830.

\bibitem[{Rethmeier et~al.(2020)Rethmeier, Kumar~Saxena, and
  Augenstein}]{pmlr-v124-rethmeier20a}
Nils Rethmeier, Vageesh Kumar~Saxena, and Isabelle Augenstein. 2020.
\newblock \href {https://proceedings.mlr.press/v124/rethmeier20a.html} {Tx-ray:
  Quantifying and explaining model-knowledge transfer in (un-)supervised nlp}.
\newblock In \emph{Proceedings of the 36th Conference on Uncertainty in
  Artificial Intelligence (UAI)}, volume 124 of \emph{Proceedings of Machine
  Learning Research}, pages 440--449. PMLR.

\bibitem[{Ribeiro et~al.(2020)Ribeiro, Wu, Guestrin, and
  Singh}]{ribeiro-etal-2020-beyond}
Marco~Tulio Ribeiro, Tongshuang Wu, Carlos Guestrin, and Sameer Singh. 2020.
\newblock \href {https://doi.org/10.18653/v1/2020.acl-main.442} {Beyond
  accuracy: Behavioral testing of {NLP} models with {C}heck{L}ist}.
\newblock In \emph{Proceedings of the 58th Annual Meeting of the Association
  for Computational Linguistics}, pages 4902--4912, Online. Association for
  Computational Linguistics.

\bibitem[{Robertson et~al.(1995)Robertson, Walker, Jones, Hancock-Beaulieu, and
  Gatford}]{robertson1995almbox}
Stephen~E Robertson, Steve Walker, Susan Jones, Micheline~M Hancock-Beaulieu,
  and Mike Gatford. 1995.
\newblock et almbox. 1995. okapi at trec-3.
\newblock \emph{Nist Special Publication Sp}, 109:109.

\bibitem[{Tenney et~al.(2020)Tenney, Wexler, Bastings, Bolukbasi, Coenen,
  Gehrmann, Jiang, Pushkarna, Radebaugh, Reif, and
  Yuan}]{tenney-etal-2020-language}
Ian Tenney, James Wexler, Jasmijn Bastings, Tolga Bolukbasi, Andy Coenen,
  Sebastian Gehrmann, Ellen Jiang, Mahima Pushkarna, Carey Radebaugh, Emily
  Reif, and Ann Yuan. 2020.
\newblock \href {https://doi.org/10.18653/v1/2020.emnlp-demos.15} {The language
  interpretability tool: Extensible, interactive visualizations and analysis
  for {NLP} models}.
\newblock In \emph{Proceedings of the 2020 Conference on Empirical Methods in
  Natural Language Processing: System Demonstrations}, pages 107--118, Online.
  Association for Computational Linguistics.

\bibitem[{Vaswani et~al.(2017)Vaswani, Shazeer, Parmar, Uszkoreit, Jones,
  Gomez, Kaiser, and Polosukhin}]{vaswani2017attention}
Ashish Vaswani, Noam Shazeer, Niki Parmar, Jakob Uszkoreit, Llion Jones,
  Aidan~N Gomez, {\L}ukasz Kaiser, and Illia Polosukhin. 2017.
\newblock Attention is all you need.
\newblock In \emph{Advances in Neural Information Processing Systems (NIPS)},
  pages 5998--6008.

\bibitem[{Vig and Belinkov(2019)}]{vig-belinkov-2019-analyzing}
Jesse Vig and Yonatan Belinkov. 2019.
\newblock \href {https://doi.org/10.18653/v1/W19-4808} {Analyzing the structure
  of attention in a transformer language model}.
\newblock In \emph{Proceedings of the 2019 ACL Workshop BlackboxNLP: Analyzing
  and Interpreting Neural Networks for NLP}, pages 63--76, Florence, Italy.
  Association for Computational Linguistics.

\bibitem[{Wallace et~al.(2019{\natexlab{a}})Wallace, Feng, Kandpal, Gardner,
  and Singh}]{wallace-etal-2019-universal}
Eric Wallace, Shi Feng, Nikhil Kandpal, Matt Gardner, and Sameer Singh.
  2019{\natexlab{a}}.
\newblock \href {https://doi.org/10.18653/v1/D19-1221} {Universal adversarial
  triggers for attacking and analyzing {NLP}}.
\newblock In \emph{Proceedings of the 2019 Conference on Empirical Methods in
  Natural Language Processing and the 9th International Joint Conference on
  Natural Language Processing (EMNLP-IJCNLP)}, pages 2153--2162, Hong Kong,
  China. Association for Computational Linguistics.

\bibitem[{Wallace et~al.(2019{\natexlab{b}})Wallace, Tuyls, Wang, Subramanian,
  Gardner, and Singh}]{wallace-etal-2019-allennlp}
Eric Wallace, Jens Tuyls, Junlin Wang, Sanjay Subramanian, Matt Gardner, and
  Sameer Singh. 2019{\natexlab{b}}.
\newblock \href {https://doi.org/10.18653/v1/D19-3002} {{A}llen{NLP} interpret:
  A framework for explaining predictions of {NLP} models}.
\newblock In \emph{Proceedings of the 2019 Conference on Empirical Methods in
  Natural Language Processing and the 9th International Joint Conference on
  Natural Language Processing (EMNLP-IJCNLP): System Demonstrations}, pages
  7--12, Hong Kong, China. Association for Computational Linguistics.

\bibitem[{Wolf et~al.(2020)Wolf, Debut, Sanh, Chaumond, Delangue, Moi, Cistac,
  Rault, Louf, Funtowicz, Davison, Shleifer, von Platen, Ma, Jernite, Plu, Xu,
  Le~Scao, Gugger, Drame, Lhoest, and Rush}]{wolf-etal-2020-transformers}
Thomas Wolf, Lysandre Debut, Victor Sanh, Julien Chaumond, Clement Delangue,
  Anthony Moi, Pierric Cistac, Tim Rault, Remi Louf, Morgan Funtowicz, Joe
  Davison, Sam Shleifer, Patrick von Platen, Clara Ma, Yacine Jernite, Julien
  Plu, Canwen Xu, Teven Le~Scao, Sylvain Gugger, Mariama Drame, Quentin Lhoest,
  and Alexander Rush. 2020.
\newblock \href {https://doi.org/10.18653/v1/2020.emnlp-demos.6} {Transformers:
  State-of-the-art natural language processing}.
\newblock In \emph{Proceedings of the 2020 Conference on Empirical Methods in
  Natural Language Processing: System Demonstrations}, pages 38--45, Online.
  Association for Computational Linguistics.

\end{thebibliography}
